\documentclass[pdflatex,sn-mathphys-num]{sn-jnl}% Math and Physical Sciences Numbered Reference Style
\usepackage{array}
\usepackage{graphicx}%
\usepackage{multirow}%
\usepackage{amsmath,amssymb,amsfonts}%
\usepackage{amsthm}%
\usepackage{mathrsfs}%
\usepackage[title]{appendix}%
\usepackage{xcolor}%
\usepackage{textcomp}%
\usepackage{manyfoot}%
\usepackage{booktabs}%
\usepackage{algorithm}%
\usepackage{algorithmicx}%
\usepackage{algpseudocode}%
\usepackage{listings}%
\usepackage{natbib}
\theoremstyle{thmstyleone}%
\theoremstyle{thmstyletwo}%

\theoremstyle{thmstylethree}%

\begin{document}

\title[Article Titl]{OpenHAIV: A Framework Towards Practical Open-World Learning}

\author*[1,2]{\fnm{Xiang} \sur{Xiang}}\email{xex@hust.edu.cn}
\author[1]{\fnm{Qinhao} \sur{Zhou}}
\author[1]{\fnm{Zhuo} \sur{Xu}}
\author[1]{\fnm{Jing} \sur{Ma}}
\author[1]{\fnm{Jiaxin} \sur{Dai}}
\author[1]{\fnm{Yifan} \sur{Liang}}
\author[1]{\fnm{Hanlin} \sur{Li}}
% \orgdiv{MoE Key Lab of Image Processing and Intelligent Control}\\

\affil[1]{ 
\centering
\orgname{HUST AI and Visual Learning Lab (HAIV Lab)},\\ \orgname{Huazhong University of Science and Technology (HUST)}, \\ \orgaddress{\street{1037 Luoyu Road}, \city{Wuhan}, Hubei \postcode{430074}, \country{China}}}

\affil[2]{ 
\centering
\orgname{Peng Cheng National Laboratory, \city{Shenzhen}, \country{China}}
}
% \author*[1,2]{\fnm{First} \sur{Author}}\email{iauthor@gmail.com}



% \affil*[1]{\orgdiv{Department}, \orgname{Organization}, \orgaddress{\street{Street}, \city{City}, \postcode{100190}, \state{State}, \country{Country}}}

% \affil[2]{\orgdiv{Department}, \orgname{Organization}, \orgaddress{\street{Street}, \city{City}, \postcode{10587}, \state{State}, \country{Country}}}

% \affil[3]{\orgdiv{Department}, \orgname{Organization}, \orgaddress{\street{Street}, \city{City}, \postcode{610101}, \state{State}, \country{Country}}}


\abstract{
Substantial progress has been made in various techniques for open-world recognition. Out-of-distribution (OOD) detection methods can effectively distinguish between known and unknown classes in the data, while incremental learning enables continuous model knowledge updates. However, in open-world scenarios, these approaches still face limitations. Relying solely on OOD detection does not facilitate knowledge updates in the model, and incremental fine-tuning typically requires supervised conditions, which significantly deviates from open-world settings. To address these challenges, this paper proposes OpenHAIV, a novel framework that integrates OOD detection, new class discovery, and incremental continual fine-tuning into a unified pipeline. This framework allows models to autonomously acquire and update knowledge in open-world environments. The proposed framework is available at \url{https://haiv-lab.github.io/openhaiv}.
}

\keywords{Open-World Learning, Out-of-Distribution, New Class Discovery, Class-Incremental Learning, Few-Shot Class-Incremental Learning}



\maketitle

\section{Introduction}
In recent years, with the continuous advancement of deep learning technologies, continual learning \cite{9349197, shaheen2022continual, wang2024comprehensive} has also rapidly evolved, giving rise to a significant number of impressive research works. Continual learning aims to enable models to retain existing knowledge while continuously acquiring new knowledge, typically derived from new data, under the constraint of limited or restricted access to data related to previously learned knowledge. Depending on the defined scenarios, continual learning is typically categorized into task-incremental learning, class-incremental learning \cite{zhou2024class}, and domain-incremental learning. Among these three settings, class-incremental learning has been the most extensively studied. Taking classification tasks as an example, class-incremental learning divides a dataset into multiple sessions, where the classes in different sessions do not overlap. The model is required to learn the classes of each session over time and is evaluated on all classes after each update. In contrast, task-incremental learning assumes knowledge of which session the data belongs to during evaluation, allowing the model to classify only within the corresponding task. On the other hand, in domain-incremental learning, different sessions involve data from the same set of classes but with different distributions. The model is required to correctly identify the domain of the test data while performing classification. Research targeting these settings has led to significant improvements in model performance in scenarios requiring multi-stage fine-tuning.

\begin{figure}[hbp]
    \centering
    \includegraphics[width=0.88\textwidth]{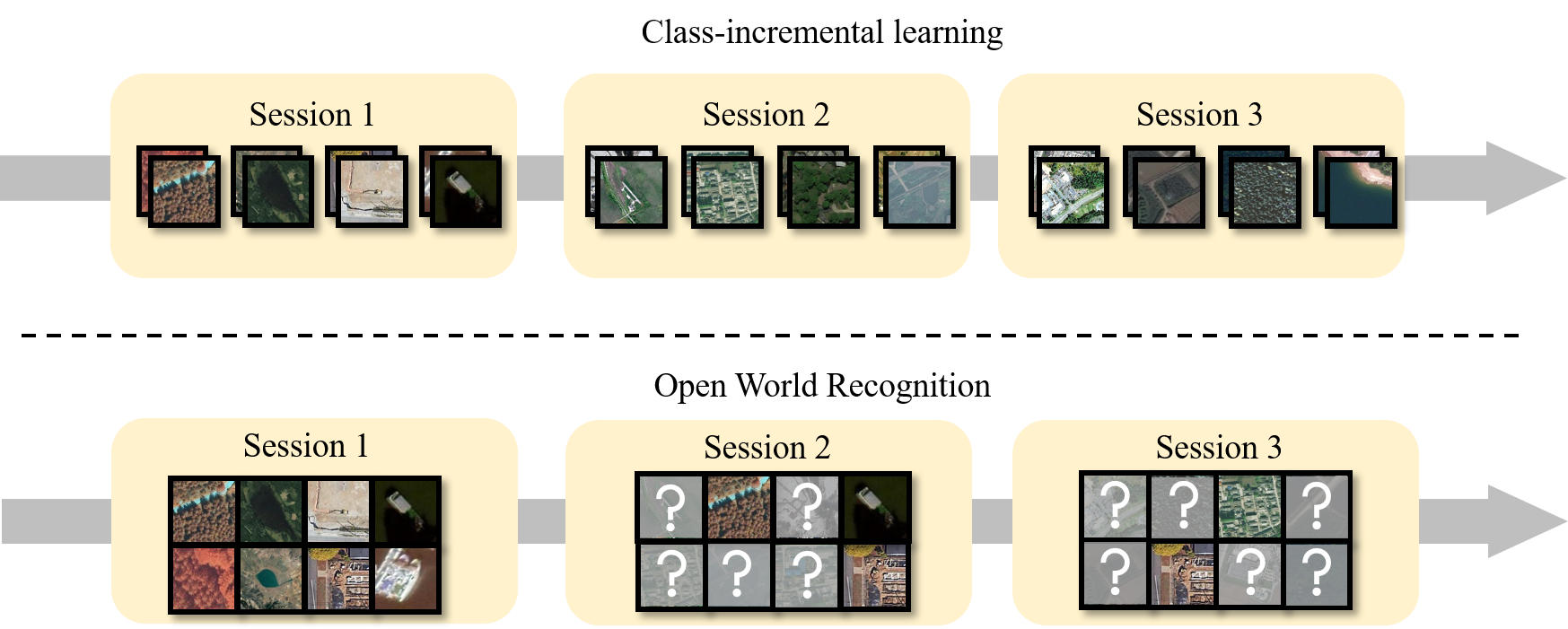}
    
    \caption{The differences between the incremental learning setting and the open-world scenario.}
    \label{intro}
\end{figure}

However, despite significant progress in incremental learning research, the current research paradigm remains confined to a closed environment, where {\bf the number of new classes in each phase and the total number of phases are predetermined}. Additionally, {\bf the data for new phases only contains new classes}, which still diverges from the realistic open-world scenarios \cite{fini2021unified, zhao2023incremental,vaze2022generalized}. In a realistic open-world setting, {\bf both the number of unknown classes and the number of phases are inherently uncertain}. Moreover, {\bf new data in each phase are likely to be a mixture of known and unknown classes}. During the knowledge updating process, the model faces various open-world challenges. When new data arrive, it must first detect which instances belong to unknown categories, a task typically addressed by out-of-distribution (OOD) detection techniques. After identifying the unknown categories, the model needs to partition them, which is commonly achieved using new class discovery algorithms. Finally, the model updates its knowledge using the newly identified unknown class data, a stage where incremental learning techniques play a primary role. Fig.~\ref{intro} illustrates the differences between the incremental learning setting and the open-world learning setting.

To address the challenge of model knowledge updating in open-world scenarios, we propose an open-world learning framework, OpenHAIV, which integrates methods from out-of-distribution detection, novel category discovery, and incremental learning. This framework establishes a comprehensive pipeline capable of enabling iterative knowledge updating for models in open-world settings. In contrast to existing frameworks focused on isolated aspects of the open-world problem, the proposed framework in this paper provides a unified architecture that simultaneously tackles multiple challenges inherent to open-world environments. By integrating diverse yet interconnected tasks into a cohesive system, this approach offers a more comprehensive and adaptive solution for real-world object learning scenarios. To validate the functionality of the proposed framework, we conducts extensive experiments on OpenEarthSensing \cite{Xiang2025OpenEarthSensingLF}, a benchmark dataset specifically designed for open-world remote sensing recognition tasks.

\section{Related Work}

% \subsection{Other Related Frameworks}
% Currently, a wide range of related frameworks have been developed, offering a diverse array of algorithms tailored to address individual problems. For instance, PyCIL \cite{zhou2023pycil} has established a class-incremental learning framework that implements multiple class-incremental learning algorithms. This framework has garnered significant attention, making class-incremental learning more accessible and explorable for the machine learning community. In addition, OpenOOD \cite{yang2022openood} implements a series of out-of-distribution detection methods. It establishes a unified benchmark for OOD tasks, offering a comprehensive comparison of various OOD methods within the framework. PILOT \cite{sun2025pilot} introduces a comprehensive framework for incremental learning based on pre-trained models, incorporating a wide range of state-of-the-art methods within a unified architecture. This framework enables systematic and fair comparisons among various pre-trained model-based incremental learning approaches, ensuring consistent evaluation metrics and experimental conditions. Mammoth \cite{} is another framework for continual learning. It implements over fifty methods and defines more than twenty datasets, utilizing a modular design that facilitates easy extensibility.

\subsection{Framework for class-incremental learning}
Class-incremental learning poses a significant challenge for machine learning models, requiring them to acquire knowledge of new classes from evolving data streams while maintaining proficiency in previously learned classes \cite{zhou2024class}. Currently, a wide range of related frameworks have been developed, offering a diverse array of algorithms tailored to address CIL problems. For instance, PyCIL \cite{zhou2023pycil} has established a class-incremental learning framework that implements multiple class-incremental learning algorithms. This framework has garnered significant attention, making class-incremental learning more accessible and explorable for the machine learning community. PyCIL primarily implements a series of conventional CIL methods trained from scratch, encompassing three major categories: data preservation methods \cite{rebuffi2017icarl, isele2018selective, rolnick2019experience, chaudhry2021using, zhao2021memory, zenke2017continual}, which mitigate catastrophic forgetting by storing and replaying exemplars from previous tasks; Regularization techniques \cite{li2017learning, kirkpatrick2017overcoming, zenke2017continual, aljundi2018memory, yang2019adaptive} that constrain parameter updates to preserve important weights for prior knowledge; parameter isolation methods \cite{mallya2018packnet, fernando2017pathnet, mallya2018piggyback}, which allocate dedicated model components for different tasks to prevent interference. Through continuous development, this framework has also incorporated relevant approaches based on pre-trained models \cite{Zhou2023RevisitingCL}. PILOT \cite{sun2025pilot} introduces a comprehensive framework for incremental learning based on pre-trained models, incorporating a wide range of state-of-the-art methods within a unified architecture. In contrast to PyCIL, this framework specifically incorporates a series of pre-trained model-based approaches, including prompt-based methods such as CODAPrompt \cite{Smith2022CODAPromptCD}, DualPrompt \cite{Wang2022DualPromptCP}, and L2P \cite{Wang2021LearningTP}. Mammoth is another framework for continual learning. It implements over fifty methods and defines more than twenty datasets, utilizing a modular design that facilitates easy extensibility. 

% Conventional approaches to class-incremental learning typically initiate model training from scratch, employing various strategies to maintain the delicate equilibrium between model stability and adaptability. These strategies can be broadly classified into three categories: data preservation methods \cite{rebuffi2017icarl, isele2018selective, rolnick2019experience, chaudhry2021using, zhao2021memory, zenke2017continual}, regularization techniques \cite{li2017learning, kirkpatrick2017overcoming, zenke2017continual, aljundi2018memory, yang2019adaptive}, and parameter isolation approaches \cite{mallya2018packnet, fernando2017pathnet, mallya2018piggyback}. These methodologies combat catastrophic forgetting through distinct yet complementary perspectives, focusing on data manipulation, model constraints, and architectural modifications. Furthermore, algorithm-centric solutions have emerged, particularly those leveraging knowledge distillation, which can be subdivided into logits-based \cite{zhang2020class, smith2021always, zhou2021co} and feature-based distillation techniques \cite{hou2019learning, lu2022augmented, park2021class}. However, models trained from scratch often lack foundational knowledge, rendering them more prone to overfitting compared to their pre-trained counterparts. This limitation is further exacerbated by the model's heavy dependence on the initial session data, resulting in inconsistent performance across different learning sessions.

\subsection{Framework for few-shot incremental learning}
Few-shot incremental learning has emerged as a significant research direction in machine learning and computer vision, attracting considerable attention from scholars worldwide. Tao et al. \cite{zhang2023few} first formalized the problem setting of Few-Shot Class-Incremental Learning (FSCIL), which presents greater challenges and demands higher generalization capabilities compared to conventional class-incremental learning. CEC \cite{zhang2021few} introduces meta-learning into FSCIL and provides an open-source implementation. This framework serves as the basis for many subsequent methods. FACT \cite{zhou2022forward} provides the first forward-compatibility analysis of few-shot incremental learning, and LIMIT \cite{zhou2022few} also successfully applies meta-learning to this field. In the base training phase, SAVC \cite{Song_2023_CVPR} incorporates supervised contrastive learning with virtual classes, while BiDist \cite{10203568} employs a dual-teacher distillation approach. These methods are developed within the CEC framework.

\subsection{ Framework for out-of-distribution detection}

OOD detection requires models to identify samples that deviate from in-distribution (ID) samples. Recently, a series of frameworks have been proposed, incorporating multiple developed methods for OOD detection tasks, while also providing convenience for subsequent algorithm development. Among them, representative frameworks include OpenOOD \cite{yang2022openood, zhang2024openood} and Dassl \cite{zhou2022domain}. OpenOOD integrates a comprehensive suite of classic unimodal OOD detection methods and provides multiple full-spectrum OOD evaluation \cite{yang2023full} benchmarks, including ImageNet \cite{deng2009imagenet} and CIFAR-100 \cite{krizhevsky2009learning}. The framework features a modular architecture that supports plug-and-play post-hoc methods \cite{hendrycks17baseline,hsu2020generalized,liu2020energy,hendrycks2019scaling,sun2021react,huang2021importance,wang2022vim}, for flexible customization, while also incorporating various training regularization-based approaches \cite{du2022vos,wei2022logitnorm}  for enhanced OOD detection performance. Recently, some work has also been proposed based on Vision-Language models (VLMs) \cite{radford2021learning, li2022blip}. An extended version of OpenOOD, named OpenOOD-VLM, has been proposed to evaluate certain VLMs-based OOD detection methods, including AdaNeg \cite{zhang2024adaneg}, LAPT \cite{zhang2024lapt}, NegLabel \cite{jiang2024negative}, MCM \cite{ming2022delving}, and others. Additionally, several fine-tuning-based methods have been developed using the Dassl framework, including LoCoOp \cite{miyai2024locoop}, SCT \cite{yu2024selfcalibratedtuningvisionlanguagemodels}, DPM \cite{2024ECCV}, and others.

\section{Proposed Framework}
We propose a comprehensive pipeline for open-world learning that integrates out-of-distribution detection, unsupervised clustering, and incremental fine-tuning methods to enable continuous knowledge updating in open-world scenarios. In this section, we will present a comprehensive exposition of the core modules and underlying design philosophy, elucidating both architectural components and theoretical foundations.

\subsection{Core Module}
\textbf{OWL:} OWL is the primary module supported by the framework, which fully implements the process of autonomous knowledge evolution for the model in an open-world setting. It enables a model \( M \) to achieve fully autonomous knowledge evolution. In the initial phase, \( M \) learns from base classes. The number of subsequent phases is uncertain, with each phase \( T_i \) having training data $ D_i = \big\{ \mathbf{x}_j \mid \mathbf{x}_j \in (\mathcal{K}_i \cup \mathcal{U}_i) \big\}_{j=1}^{N} $, where $\mathcal{K}_i$ represents known classes and $\mathcal{U}_i$ represents unknown classes, both of which are unannotated. The model needs to differentiate between the unknown classes and subsequently update its parameters using the data from these unknown classes. Similarly, during evaluation, the model \( M \) is required to be assessed on all previously encountered classes.

\textbf{OOD:} OOD is the key module in implementing OWL, facilitating the discovery of new unknown class data at each session. Given the trained model \( M \), OOD detection aims to identify whether an input sample \( x \) is drawn from a distribution different from the training data distribution or in-distribution \( P_{\text{ID}}(x) \). Formally, let \( P_{\text{test}}(x) \) be the distribution of test data. Then, \( x \) is considered OOD if \( P_{\text{test}}(x) \) significantly deviates from \( P_{\text{ID}}(x) \). The goal of OOD detection is to design a score function $S(x)$ for ID/OOD binary classification.  Recent work has categorized OOD tasks into two types: semantic OOD detection and covariate shift OOD detection. The former focuses on detecting OOD samples with semantic shifts, while the latter incorporates covariate-shifted ID data during testing, requiring the model to generalize to such data. Closely related task is open-set recognition (OSR), which also demands accurate of the ID samples.

\textbf{NCD:} This module aims to autonomously identify and characterize previously unknown categories in unlabeled data $D_u=\{x_i\}_1^M$. Conventional approaches employ unsupervised clustering techniques, such as K-Means, to partition the data based on sample features $V_u=\{v_i\}_1^M$, thereby establishing categorical differentiation without prior knowledge.

\textbf{CIL:} CIL functions as the key module for model updates within the framework. The model $M$ is divided into a feature extractor $B$ and a classifier $F$, and it undergoes continuous training. The feature extractor is shared across all sessions, while the parameters of the classification head progressively increase with each subsequent session. For standard CIL setting, the training regimen is segmented into \( T \) distinct sessions, each dedicated to training a specific subset of classes, denoted collectively as \( C = \{C_1, C_2, \ldots, C_T\} \). Importantly, the class subsets across sessions are mutually exclusive, meaning for any two sessions \( i \) and \( j \) where \( i \neq j \), the intersection \( C_i \cap C_j \) is empty. For each session \( t \), a training dataset \( X_t \) is established, consisting of pairs \( \{(x_1, y_1), (x_2, y_2), \ldots, (x_N, y_N)\} \) corresponding to the categories in \( C_t \). The model \( M \) is restricted to fine-tuning exclusively on the current session's training set. During evaluation, the model is required to accurately classify instances from all categories it has encountered thus far. This setup is designed to ensure robust recognition performance across the entirety of learned categories.

\textbf{FSCIL:} Similar to CIL, FSCIL also partitions the data into $T$ incremental sessions. However, a key distinction lies in the learning paradigm: in FSCIL, the base session first trains the model on a large number of classes, while each subsequent incremental session introduces $N$ novel classes with only $K$ samples per class, $K$ is very small (e.g., $1–5$). This setting is commonly referred to as N-way K-shot few-shot learning, posing significant challenges due to the extreme scarcity of training data in later sessions.

\begin{figure}[hbp]
    \centering
    \includegraphics[width=0.98\textwidth]{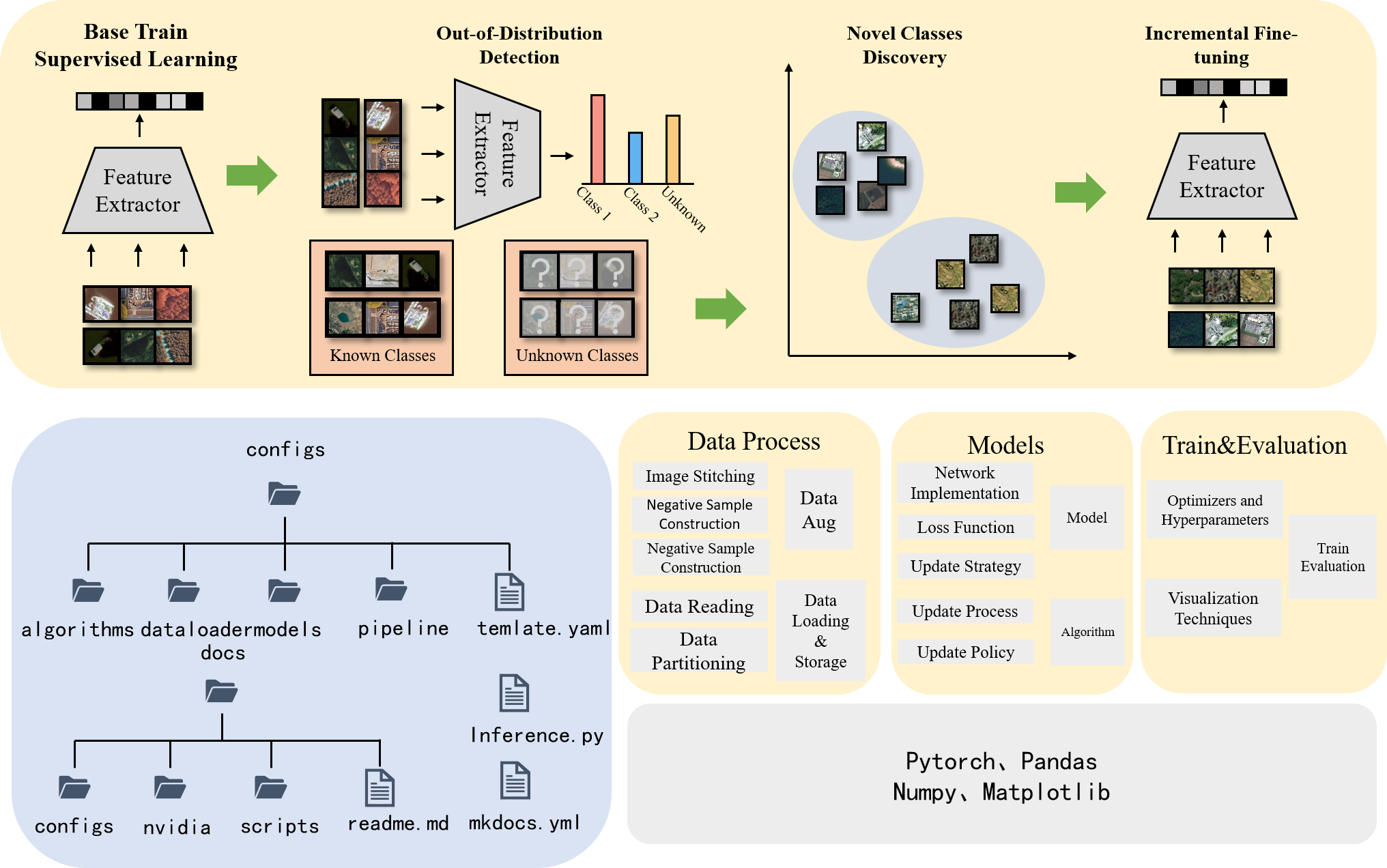}
    
    \caption{The pipeline and structure of our framework. \textbf{Top:} the pipeline for open-world learning. \textbf{Bottom:} The files structure and modules of the framework.}
    \label{pipeline}
\end{figure}

\subsection{Framework Design}

The framework adopts a modular design overall, which is reflected in two key aspects. Functionally, it independently incorporates dedicated modules for supervised training, out-of-distribution detection, novel class discovery, and incremental learning. Procedurally, we divide the operational workflow into distinct stages, including data processing, model construction, training \& evaluation, and visualization. From an implementation perspective, the framework is built upon foundational deep learning, data processing, and visualization libraries such as PyTorch, NumPy, Matplotlib, and Pandas, leveraging their extensive built-in functionalities. Fig~\ref{pipeline} shows the pipeline and every modular of the framework.

The data processing module comprises two key components: data loading and data augmentation. The data loading function primarily converts image data into multi-dimensional matrices for model input, while also performing dataset partitioning that not only separates training and testing sets but also divides data into multiple phases for incremental learning, with each phase containing a specific number of class categories to facilitate progressive model adaptation. The data augmentation component implements various image preprocessing techniques including image stitching, negative sample construction, rotation, cropping, and format conversion to enhance model robustness.

The model construction module integrates two key components: network architectures and algorithmic strategies. The architecture component provides configurable deep neural networks with diverse loss functions, enabling task-specific customization while critically influencing training dynamics. The algorithmic component implements flexible update mechanisms across three specialized submodules: (1) out-of-distribution detection combining prediction-based and feature-based methods, (2) novel class discovery through advanced clustering techniques, and (3) incremental learning incorporating knowledge distillation, regularization, and few-shot learning approaches to address catastrophic forgetting. This unified design ensures adaptive model evolution while maintaining backward compatibility across learning phases.

The training and evaluation module supports multiple optimizers and enables flexible hyperparameter configuration, providing essential tools for optimizing model training processes. This design allows users to adapt optimization strategies according to specific task requirements and model characteristics, ensuring optimal training performance. By offering such flexibility, users can select the most suitable optimization algorithms and hyperparameter settings tailored to different learning scenarios and architectural demands.

The framework exhibits excellent extensibility by adopting a modular design, which facilitates development and future functional expansion. Each module operates independently, interacting through well-defined interfaces, thereby ensuring high flexibility in internal development. This modular architecture not only enhances code readability and maintainability but also allows for isolated modifications and optimizations of specific modules without affecting others. Moreover, the independence between modules significantly simplifies the integration of new functionalities, enabling seamless addition of new modules or replacement of existing ones during development.

\subsection{Basic Usage}
\textbf{OOD.} The OpenHAIV framework integrates more than $10$ distinct OOD detection algorithms, along with standardized evaluation protocols on the OpenEarthSensing benchmark under various OOD scenarios, facilitating consistent and reliable comparisons across different methods. Users can easily customize their evaluation settings, allowing for tailored experimental designs to meet their needs. 

\textbf{CIL and FSCIL.} OpenHAIV integrates $10$ CIL algorithms and FSCIL algorithms, offering standardized evaluation protocols for OpenEarthSensing and other natural image datasets under class-incremental learning setting. The framework is modularly designed to support rapid adoption of new datasets and customized incremental learning strategies. We provides granular control over hyperparameter configurations, offering diverse customizable options for each training module to accommodate varied experimental requirements. The framework provides flexible mechanisms for extending core components including methods, networks, and datasets. Users can seamlessly register new modules through standardized interfaces. Additionally, a hook-based mechanism enables the injection of custom operations during the training pipeline, offering fine-grained control over the learning process.

\section{Experiments}

\subsection{Datasets and Evaluation Protocols}
We conduct experiments on the OpenEarthSensing dataset, a remote sensing recognition dataset specifically designed for open-world identification. The dataset consists of $157,674$ images, organized into 10 broad categories and 189 fine-grained subcategories, covering both scene-level and object-level classifications, and includes images of diverse scales. The dataset supports various open-world tasks, including out-of-distribution detection and incremental learning. We utilize this dataset to simulate open-world learning settings and conduct corresponding experiments. Additionally, we perform experiments on various subtasks within the open-world learning domain. 
For OWL, in the base session, the model learns $94$ classes, followed by a subsequent session with $95$ classes. The training data in this latter session is unlabeled and includes both known and unknown classes. Ultimately, the model aims to learn all $189$ classes.
For CIL, we evaluate two different orders: random, coarse-to-fine based order. In the random order experiments, we follow the conventional CIL setting, where where each session involves learning a random set of $20$ classes, totaling $10$ sessions. In the coarse-to-fine order experiments, we also employed the standard CIL setup, requiring the category order to progress from coarse classes to fine classes. For OOD, we utilize RGB images from $94$ classes as ID training data, and evaluate the standard and four covariate-shift OOD setting on OES benchmark, including resampling bias, aerial modal shift, MSRGB modal shift and IR modal shift.
For FSCIL, we one order. Initial session includes 109 classes and each incremental session includes $10$ classes, which is $10$-way-$5$-shot learning paradigm.

% \usepackage{multirow}
\begin{table}[h]
\caption{The experiments of OWL on OpenEarthSensing dataset. ID Acc and ODD Acc are the in-distribution and out-of-distribution performance, respectively, and Avg denotes the average performance of each session.}
\begin{tabular}{ccccccc} \toprule
OOD   Method         & CIL Method & ID Acc            & OOD Acc           & Session 1              & Session 2            & Avg                  \\ \hline
\multirow{3}{*}{MSP \cite{hendrycks17baseline}} & LwF \cite{li2017learning}& \multirow{3}{*}{91.17} & \multirow{3}{*}{55.01} & \multirow{3}{*}{91.27} & 42.11                & 66.69                \\
                     & EWC \cite{kirkpatrick2017overcoming}&                   &                   &                        & 28.89                & 60.08                     \\
                     & iCaRL \cite{rebuffi2017icarl}&                   &                   &                        & 50.29                & 70.78                     \\ \hline
\multirow{3}{*}{MLS \cite{species22icml}} & LwF \cite{li2017learning}& \multirow{3}{*}{90.6} & \multirow{3}{*}{63.85} & \multirow{3}{*}{91.27} &    44.09 &    67.68\\ 
                     & EWC \cite{kirkpatrick2017overcoming}&                   &                   &                        &     29.83             &  60.55                     \\
                     & iCaRL \cite{rebuffi2017icarl}&                   &                   &                        &   51.33  &     71.30                     \\ \hline
\multirow{3}{*}{VIM \cite{wang2022vim}} & LwF \cite{li2017learning}& \multirow{3}{*}{93.5} & \multirow{3}{*}{59.99} & \multirow{3}{*}{91.27} & 43.49& 67.38\\
                     & EWC \cite{kirkpatrick2017overcoming}&                   &                   &                        & 31.33 &     61.30                     \\
                     & iCaRL \cite{rebuffi2017icarl}&                   &                   &                        & 33.78 &     62.52                     \\
                     \bottomrule
\end{tabular}
\label{owl}
\end{table}

\begin{figure*}
\hsize=\textwidth 
\centering
\includegraphics[width=0.92\textwidth]{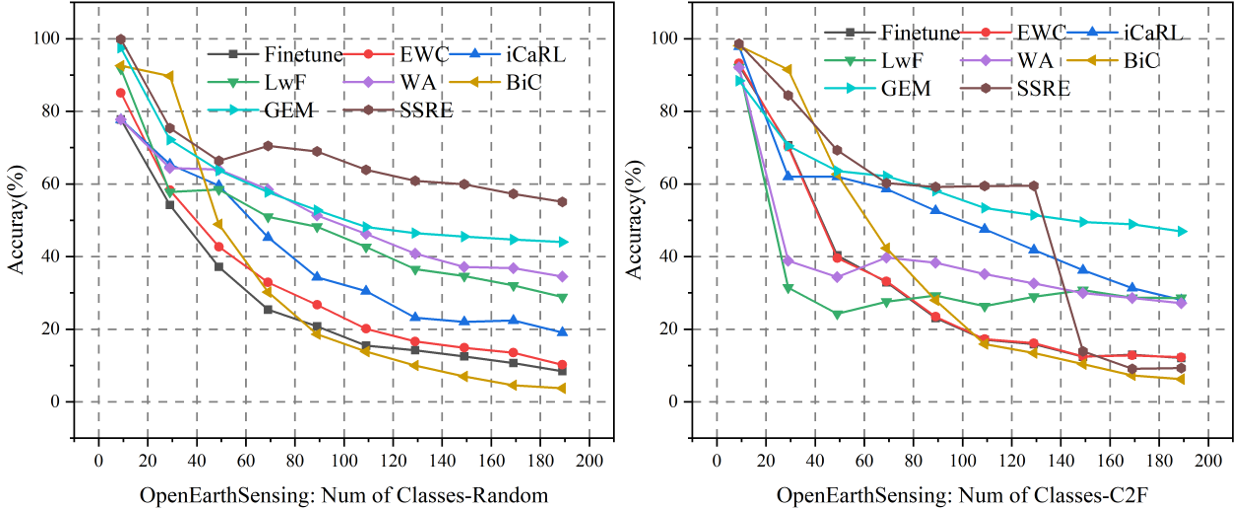}

\caption{ Experimental results of CIL. The left figure presents results in randomized order, while the right figure displays systematically organized results arranged from coarse to fine granularity.
}

\label{cil}
\end{figure*}

\subsection{Results}

We select commonly used methods in OOD and CIL for OWL experiments and the results of OWL can be seen in tab.~\ref{owl}. It can be observed that during the base session, all methods achieve satisfactory performance. However, in the subsequent session, the inability to ensure that the new classes are completely free of noise leads to a significant decline in performance across all methods.

\begin{table}[h]
\centering
\caption{The experimental results of few-shot class-incremental learning on the OpenEarthSensing dataset. Shots denote the training samples for each category.}
\begin{tabular}{ccccccccc} \toprule
      & \multicolumn{2}{c}{50-shot} & \multicolumn{2}{c}{10-shot} & \multicolumn{2}{c}{5-shot} & \multicolumn{2}{c}{1-shot} \\ \hline
      & Last         & Avg          & Last         & Avg          & Last         & Avg         & Last         & Avg         \\
Alice \cite{Peng2022FewShotCL}& 59.54        & 64.66        & 59.17        & 68.64        & 58.82        & 68.35       & 58.94        & 68.4        \\
FACT \cite{Zhou2022ForwardCF}& 46.42        & 49.21        & 46.38        & 49.15        & 46.36        & 49.15       &    46.37          & 49.15            \\
SAVC \cite{Song2023LearningWF}&  71.71            &  79.55            &  72.23            &   80.07  & 66.61 & 75.17 &    59.63          & 76.77     \\ \bottomrule      
\end{tabular}
\label{fscil}
\end{table}

We conduct extensive CIL experiments using a ResNet-18 architecture to evaluate eight representative methods: Finetune as the baseline approach, followed by EWC \cite{kirkpatrick2017overcoming}, iCaRL \cite{rebuffi2017icarl}, LwF \cite{li2017learning}, WA \cite{zhao2020maintaining}, BiC \cite{wu2019large}, GEM \cite{lopez2017gradient}, and SSRE \cite{zhu2022self}. This systematic comparison provides insights into the relative strengths of different incremental learning strategies under standardized conditions. The results of CIL can be seen in fig.~\ref{cil}. It can be seen that traditional CIL methods show limited performance on OpenEarthSensing, with severe catastrophic forgetting persisting. Although replay-based and architecture-expanding approaches work well on standard benchmarks, this dataset's varied scales, resolutions and class distributions better match real-world complexity.

\begin{table}
\caption{ OOD detection performance on OES benchmark. 'Near' represents the average AUROC for Near-OOD datasets, 'Far' indicates the average AUROC for Far-OOD datasets.}

\setlength{\tabcolsep}{1.8mm}{
 % Change to \footnotesize for a smaller font
\begin{tabular}{c| cc|cc  |cc |cc|cc}  \toprule
\multirow{2}{*}{Method} & \multicolumn{2}{c}{Standard} & \multicolumn{2}{c}{Res Bias} & \multicolumn{2}{c}{Aerial} & \multicolumn{2}{c}{MSRGB} & \multicolumn{2}{c}{IR} \\
\cline{2-11}
& Near & Far &  Near & Far  &  Near & Far &  Near & Far  &  Near & Far   \\
\hline
% OpenMax \cite{openmax16cvpr}& 84.12 & 84.4 & 92.01 & 55.02 & 47.3 & 47.73 & 48.14 & 39.71 & 18.41 & 50.98 & 52.45 & 46.59 & 49.98 & 44.10 & 31.06 \\

MSP \cite{hendrycks17baseline} & 88.42 & 93.91 & 66.51 & 78.4 & 54.38 & 56.85 & 65.50 & 66.92 & 61.47 & 65.35 \\

ODIN \cite{odin18iclr} & 87.14 & 95.79 & 67.09 & 84.2 & 52.85 & 57.04 & 66.55 & 61.75 & 62.11 & 73.28 \\
 
MDS \cite{mahananobis18nips} & 83.51 & 96.54 & 53.86 & 75.71 & 48.79 & 54.76 & 56.31 & 85.58 & 83.64 & 57.74 \\

MLS \cite{species22icml} & 88.59 & 96.12 & 66.44 & 83.17 & 53.93 & 59.78 & 64.46 & 63.37 & 62.49 & 67.06 \\

VIM \cite{wang2022vim}  & 90.35 & 98.75 & 60.33 & 83.93 & 50.69 & 59.72 & 59.40 & 81.75 & 57.65 & 51.08 \\

FBDB \cite{liu2024fast} & 90.24 & 98.17 & 66.64 & 87.87 & 54.49 & 60.41 & 64.95 & 74.49 & 61.40 & 68.62 \\

VOS \cite{vos22iclr} & 86.19 & 95.68 & 63.37 & 81.32 & 51.10 & 60.01 & 59.77 & 58.72 & 59.47 & 60.26 \\

LogitNorm \cite{wei2022mitigating} & 89.00 & 95.15 & 68.80 & 80.29 & 53.25 & 56.72 & 77.69 & 55.43 & 64.12 & 63.97 \\

DML \cite{Zhang_2023_CVPR} &  84.38 & 90.36 & 65.78 & 76.16 & 52.89 & 58.60 & 62.56 & 50.68 & 60.39 & 50.56 \\

\bottomrule
\end{tabular}
 % Change to \footnotesize for a smaller font
}

\label{overview of ood}
\end{table}

We also conduct experiments on FSCIL setting using a ResNet-18 architecture to evaluate three representative methods: Alice, FACT and SAVC. Experiments are conducted under multiple shot configurations, and the results are presented in tab.~\ref{fscil}. The results demonstrate significant performance fluctuations across methods in the few-shot setting. and SAVC achieves highest performance among three methods.

\begin{table}
\caption{ CLIP based methods' performance on OES benchmark. 'Near' represents the average AUROC for Near-OOD datasets, 'Far' indicates the average AUROC for Far-OOD datasets
}

\setlength{\tabcolsep}{1.8mm}{
 % Change to \footnotesize for a smaller font
\begin{tabular}{c| cc|cc  |cc |cc|cc}  \toprule
\multirow{2}{*}{Method} & \multicolumn{2}{c}{Standard} & \multicolumn{2}{c}{Res Bias} & \multicolumn{2}{c}{Aerial} & \multicolumn{2}{c}{MSRGB} & \multicolumn{2}{c}{IR} \\
\cline{2-11}
& Near & Far &  Near & Far  &  Near & Far &  Near & Far  &  Near & Far   \\
\hline
% OpenMax \cite{openmax16cvpr}& 84.12 & 84.4 & 92.01 & 55.02 & 47.3 & 47.73 & 48.14 & 39.71 & 18.41 & 50.98 & 52.45 & 46.59 & 49.98 & 44.10 & 31.06 \\

MaxLogits \cite{species22icml} & 53.31 & 43.95 & 68.99 & 63.32 & 64.73 & 40.46 & 68.22 & 9.34 & 62.73 & 37.00 \\

MCM  \cite{ming2022delving} & 61.79 & 52.60 & 59.07 & 51.94 & 66.07 & 67.85 & 58.90 & 55.89 & 54.41 & 40.43 \\
 
GLMCM \cite{miyai2025zero}& 62.07 & 52.33 & 59.29 & 51.57 & 65.20 & 67.42 & 57.32 & 56.89 & 54.75 & 42.30 \\

CoOp \cite{zhou2022learning}& 86.04 & 94.21 & 64.09 & 73.36 & 61.22 & 76.40 & 66.73 & 90.22 & 61.30 & 45.16 \\

LoCoOp \cite{miyai2024locoop}& 85.71 & 90.94 & 66.20 & 71.67 & 64.18 & 76.52 & 69.64 & 86.28 & 61.41 & 43.33 \\

SCT \cite{yu2024selfcalibratedtuningvisionlanguagemodels} & 85.56 & 90.78 & 65.37 & 70.30 & 64.14 & 77.67 & 68.58 & 86.41 & 60.81 & 42.48 \\

DPM \cite{2024ECCV}  & 91.19 & 99.24 & 68.60 & 92.61 & 60.50 & 71.26 & 74.66 & 93.56 & 65.11 & 75.10 \\

\bottomrule
\end{tabular}
 % Change to \footnotesize for a smaller font
}

\label{clip ood}
\end{table}

We evaluate both conventional unimodal-based methods and vision-language model-based approaches for OOD detection on the OES benchmark. For unimodal-based methods, we use the ResNet-50 architecture to assess the following representative methods: MSP \cite{hendrycks17baseline}, MLS \cite{species22icml}, ODIN \cite{odin18iclr}, VIM  \cite{wang2022vim}, MDS  \cite{mahananobis18nips}, FBDB  \cite{liu2024fast}, LogitNorm  \cite{wei2022mitigating}, VOS \cite{vos22iclr} and DML  \cite{Zhang_2023_CVPR}. 
Considering the absence of directly applicable pretrained models, all post-hoc methods are evaluated using a ResNet-50 model trained with cross-entropy loss on ID training data for 100 epochs with a 0.05 learning rate. All training-required methods are further fine-tuned for 30 additional epochs with a learning rate of 0.01. Results across 5 OOD settings are presented in fig.~\ref{overview of ood}. For CLIP-based OOD detection methods, we use the remote sensing pretrained GeoRSCLIP \cite{zhang2024rs5m}, with ViT-b32 architecture to evaluate the following methods: Maxlogit \cite{species22icml}, MCM \cite{ming2022delving}, GLMCM \cite{miyai2025zero}, CoOp \cite{zhou2022learning}, LoCoOp \cite{miyai2024locoop}, SCT \cite{yu2024selfcalibratedtuningvisionlanguagemodels}, DPM \cite{2024ECCV}. All the training required methods are fine-tuned for 20 epochs with a 0.01 learning rate. The results of 5 OOD settings are shown in Fig.~\ref{clip ood}.

\section{Conclusion}\label{sec13}

\backmatter

This study systematically investigates the critical challenges in open-world learning, where existing approaches typically address isolated aspects of the problem. To bridge this gap, we propose a novel framework that harmoniously integrates three key capabilities: out-of-distribution detection for identifying unknown instances, novel class discovery for autonomously recognizing emerging categories, and incremental learning for continuous knowledge integration. Through comprehensive experiments on the OpenEarthSensing benchmark, we validate the functional efficacy of the proposed framework. 

% need ?
% \section{Declarations}

% \bmhead{Acknowledgements}

% Acknowledgements are not compulsory. Where included they should be brief. Grant or contribution numbers may be acknowledged.

% Please refer to Journal-level guidance for any specific requirements.

% \bibliographystyle{sn-basic} % 设置参考文献样式
\bibliography{sn-bibliography} % 指定参考文献文件

\end{document}